\documentclass[letterpaper, 10 pt, conference]{ieeeconf}  

\IEEEoverridecommandlockouts                              

\usepackage{amsmath,amssymb,balance,booktabs,cite,changes,color,enumerate,float,graphicx,hyperref,multirow,siunitx}
\usepackage[font=footnotesize]{caption}
\newcommand{\alert}[1]{\textbf{}}

\newcommand{\BEAS}{\begin{eqnarray*}}
\newcommand{\EEAS}{\end{eqnarray*}}
\newcommand{\BEA}{\begin{eqnarray}}
\newcommand{\EEA}{\end{eqnarray}}
\newcommand{\BEQ}{\begin{equation}}
\newcommand{\EEQ}{\end{equation}}
\newcommand{\BIT}{\begin{itemize}}
\newcommand{\EIT}{\end{itemize}}
\newcommand{\BNUM}{\begin{enumerate}}
\newcommand{\ENUM}{\end{enumerate}}
\newcommand{\BEL}[1]{\begin{equation}\label{#1}}
\newcommand{\EEL}{\end{equation}}

\newcommand{\state}{\mathbf{s}}
\newcommand{\goal}{\mathbf{g}}
\newcommand{\traj}{\tau}

\newcommand{\action}{\mathbf{a}}

\newcommand{\policy}{\pi}

\newcommand{\buff}{\mathcal{D}}
\newcommand{\nrefs}{{N_\text{refs}}}

\newcommand{\BA}{\begin{array}}
\newcommand{\EA}{\end{array}}

\usepackage{algorithm}
\usepackage[noend]{algpseudocode}

\definecolor{blue}{rgb}{0.44, 0.65, 0.82}
\definechangesauthor[color=blue, name=laura]{LS}

\title{\LARGE \bf Legged Robots that Keep on Learning:\\ Fine-Tuning Locomotion Policies in the Real World}

\author{
\authorblockN{Laura Smith\textsuperscript{1}, J. Chase Kew\textsuperscript{2}, Xue Bin Peng\textsuperscript{1}, Sehoon Ha\textsuperscript{2,3}, Jie Tan\textsuperscript{2}, Sergey Levine\textsuperscript{1,2}}
\authorblockA{\textsuperscript{1}Berkeley AI Research, UC Berkeley \textsuperscript{2}Google Research \textsuperscript{3}Georgia Institute of Technology\\
Email: \texttt{smithlaura@berkeley.edu}}
}

\begin{document}
\setlength{\textfloatsep}{7pt}

\makeatletter
\let\@oldmaketitle\@maketitle%
\renewcommand{\@maketitle}{\@oldmaketitle%
    \centering
    \includegraphics[width=1.0\linewidth]{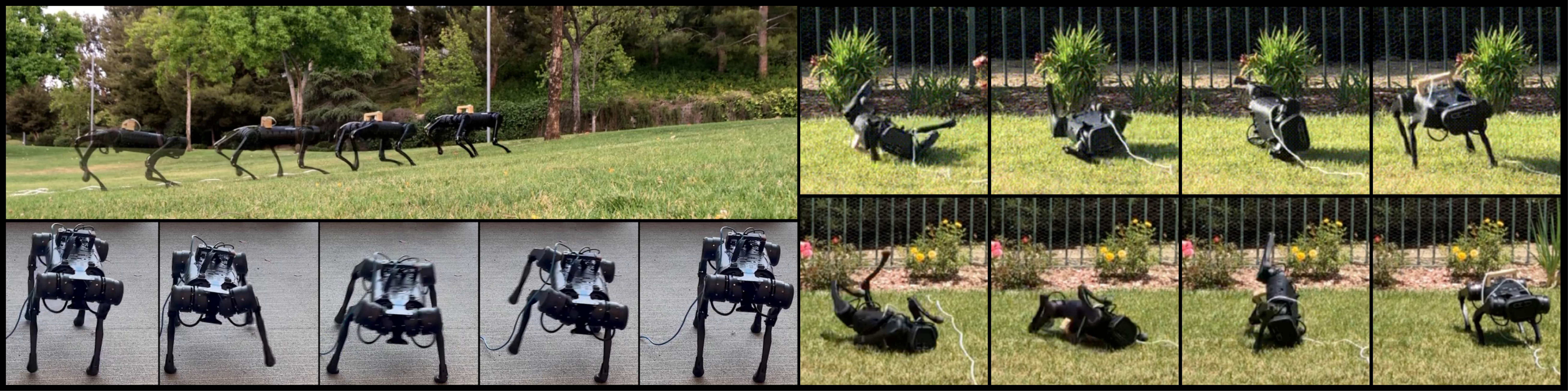}
    \captionof{figure}{\footnotesize We demonstrate real-world improvement through fine-tuning multiple skills to various real-world environments. The robot learns to walk back and forth on grass (\textbf{top left}) and side-step on carpet (\textbf{bottom left}), while recovering seamlessly from failure (\textbf{right}).}
    \vspace{-.4cm}
    \label{fig:teaser}
}
\makeatother

\maketitle
\thispagestyle{empty}
\pagestyle{empty}

\begin{abstract}
Legged robots are physically capable of traversing a wide range of challenging environments, but designing controllers that are sufficiently robust to handle this diversity has been a long-standing challenge in robotics. Reinforcement learning presents an appealing approach for automating the controller design process and has been able to produce remarkably robust controllers when trained in a suitable range of environments. However, it is difficult to predict all likely conditions the robot will encounter during deployment and enumerate them at training-time. What if instead of training controllers that are robust enough to handle any eventuality, we enable the robot to continually learn in any setting it finds itself in?
This kind of real-world reinforcement learning poses a number of challenges, including efficiency, safety, and autonomy. To address these challenges, we propose a practical robot reinforcement learning system for fine-tuning locomotion policies in the real world. We demonstrate that a modest amount of real-world training can substantially improve performance during deployment, and this enables a real A1 quadrupedal robot to autonomously fine-tune multiple locomotion skills in a range of environments, including an outdoor lawn and a variety of indoor terrains. (Videos and code\footnote{\tt \url{https://sites.google.com/berkeley.edu/fine-tuning-locomotion}})
\end{abstract}


\section{Introduction}
\label{sec:intro}
Legged robots possess a unique physical capability to traverse a wide range of environments and terrains, from subterranean rubble to snowy hills~\cite{Miller2020MineTE, Lee2020LearningQL}. However, fully realizing this capability requires controllers that can effectively handle this broad range of environments. Engineering such robust controllers for each robot is a labor-intensive process, requiring human expertise and precise modeling of the system dynamics~\cite{Bledt2018MITC3, Gehring2016PracticeMP, Hutter2016ANYmalA}. Reinforcement learning (RL) algorithms have been used to automatically learn robotic locomotion skills in a wide range of contexts, both in simulation and in the real world~\cite{Kohl2004PolicyGR, Tedrake2004StochasticPG, Tedrake2005LearningTW, Endo2005LearningCS, Heess2017EmergenceOL, Tan2018SimtoRealLA, Xie2018FeedbackCF, Liu2018LearningBD, Peng2018DeepMimicED, Lee2019ScalableMH, Haarnoja2019LearningTW,  Hwangbo2019LearningAA}.
However, in order for these methods to handle the full range of environments that the robot will encounter at test-time, they must be trained in an appropriately broad range of conditions -- in a sense, the RL approach exchanges the burden of controller engineering for the burden of training-time environment engineering. While much of the work on learning locomotion skills has focused on training robust skills that can generalize to a variety of test-time conditions (e.g., different terrains)~\cite{Kumar2021RMA, Peng2020LearningAR}, they all share the same fundamental limitation: they lack any recourse when the test-time conditions are so different that the trained controllers fail to generalize.
In this paper we are specifically interested in the case where perfect zero-shot generalization is impossible. 
In this case, what can the robot do? In this paper, our approach to this problem is to enable \emph{real-world fine-tuning}: when the robot inevitably fails, it would need the mechanisms necessary to recover and \emph{fine-tune} its skills to this new environment.

Although in principle RL provides precisely the toolkit needed for this type of adaptation, in practice this kind of fine-tuning presents a number of major challenges: the fine-tuning must be performed rapidly, under real-world conditions, without reliance on external state estimation or human assistance. The goal in this paper is to design a complete system for fine-tuning robotic locomotion policies under such real-world conditions. In our proposed framework, the robot would first attempt the desired locomotion task in some new environment, such as the park shown in~\autoref{fig:teaser}. Initially, it may fall because the uneven ground is not compatible with its learned policy. At this point, it should immediately stand back up using an agile learned reset controller, make a few more attempts at the task, and then use the collected experience to update its policy. 
To enable open world learning, the reward signal for RL must be obtained from the robot's own onboard sensors, and the robot must keep attempting the task until it succeeds, improving with each trial. This process must be successful both where it generalizes well (and is usually successful) and where it generalizes poorly (and fails on most initial trials).
Concretely, we utilize motion imitation~\cite{Peng2020LearningAR} to provide a general recipe for learning agile behaviors. To ensure that the robot operates autonomously, we use a learned recovery policy that enables the robot to quickly and robustly recover from falls. This autonomous process can either fine-tune one skill at a time, or fine-tune multiple complementary skills together, such as a forward and a backward walking motion. Due to our choice of RL formulation, learning these additional policies is a simple, straightforward extension.
For efficient and stable real-world training, we opt for an off-policy RL algorithm that uses randomization over ensembles to stabilize and substantially improve the sample-efficiency of Q-learning methods~\cite{Chen2021RandomizedED}.

The main contribution of our work is a system for real-world autonomous fine-tuning of agile quadrupedal locomotion skills. To our knowledge, our system is the first to show real-world fine-tuning using RL, with automated resets and onboard state estimation, for multiple agile behaviors with an underactuated robot.
In our experiments, we take advantage of simulation data to pre-train a policy, reaping the safety and efficiency benefits of training in simulation, while retaining the ability to continue learning in new environments with real-world training. 
Although the particular components we integrate into our real-world fine-tuning system are based on prior works, the combination of these components is unique to our system and together, they enable efficient real-world fine-tuning of agile and varied locomotion behaviors, together with highly efficient resets between trials and recoveries from falls. 
We demonstrate in our experiments that our system enables an A1 quadruped robot to learn dynamic skills, such as pacing forwards and backwards in an outdoor grass field, and side-stepping on 3 indoor terrains: carpet, doormat with crevices, and memory foam.

\vspace{-.15cm} 
\section{Related Work}
\label{sec:relatedwork}
Robotic locomotion controllers are typically constructed via a combination of footstep planning, trajectory optimization, and model-predictive control (MPC)~\cite{Katz2019MiniCA}. This has enabled a range of desirable gaits, including robust walking~\cite{Hutter2016ANYmalA} and high-speed bounding~\cite{Park2017BoundingCheetah}. However, such methods require characterization of the robot's dynamics and typically a considerable amount of manual design for each robot and each behavior. RL provides an appealing approach to instead learn such skills, both in simulation~\cite{Liu2018LearningBD, Lee2019ScalableMH, Peng2018DeepMimicED} and in the real world~\cite{Kohl2004PolicyGR, Tedrake2004StochasticPG, Endo2005LearningCS, Tan2018SimtoRealLA, Haarnoja2019LearningTW, Hwangbo2019LearningAA}. Due to safety considerations and the data intensive nature of RL algorithms, RL-based locomotion controllers are often trained in simulation. Various methods are used to improve transfer to the real world, such as building high fidelity simulators~\cite{Tan2018SimtoRealLA, Xie2019LearningLS}, using real world data to improve the accuracy of simulators~\cite{Tan2016SimulationbasedDO, Chebotar2019ClosingTS, Du2021AutoTunedST}, and simulating diverse conditions
to capture the variations a robot may encounter during real world deployment~\cite{Peng2018SimtoRealTO, sadeghi2016cad2rl, Lee2020LearningQL}. However, legged robots are capable to traverse such a wide variety of terrains. It is thus difficult to anticipate all the conditions they may encounter at test-time, and even the most robust learned policies may not generalize to every such situation. 

Another line of work trains adaptive policies by incorporating various domain adaptation techniques to perform few-shot adaptation~\cite{He2018ZeroShotSC, Yu2019SimtoRealTF, Hwangbo2019LearningAA, Xie2019LearningLS, Yu2020LearningFA, Peng2020LearningAR}. Particularly related to our work, two prior works have proposed to train locomotion policies in simulation~\cite{Peng2020LearningAR, Kumar2021RMA} that include a learned adaptation structure, which infers a latent or explicit descriptor of the environment. However, although such policies are adaptive, their ability to adapt is also limited by the variability of conditions seen at training-time --- if the test-time conditions differ in ways that the designer of the simulation did not anticipate, they may likewise fail, as we illustrate in our experimental comparison in~\autoref{sec:simulation-experiments}.
Thus, in this work, we rather aim to perform consistent adaptation through fine-tuning with RL, and present a method that enables continuous improvement under any test-time condition in the real world. Julian et al. \cite{Julian2020EfficientAF} uses an off-policy model-free RL approach to fine-tune a visual grasping policy to a variety of conditions that are not covered during pre-training. Our approach similarly uses off-policy model-free learning to continuously learn subject to changes in the environment, but we instead consider a variety of skills and challenges introduced by learning legged locomotion skills, such as underactuation and falling.

Several works have approached the challenge of training locomotion policies in the real world. This approach, however, has yet to scale to more complex motions due to the supervision requirements: these systems often rely on heavy instrumentation of the environment, such as motion capture systems, to provide reward supervision and resets, through engineering~\cite{Ha2020LearningTW} or manual human intervention~\cite{Haarnoja2019LearningTW, Yang2019DataER, Choi2019TrajectorybasedPP}. To make real-world training more broadly applicable (e.g., in the outdoors) we perform all state estimation on board, without any motion capture or external perception. While these prior works have demonstrated learning very conservative walking gaits on simple robots in the real world from scratch, we demonstrate learning of pacing and side-stepping, behaviors that are naturally unstable and require careful balancing on a more agile A1 robot. Thus, we found that it was crucial to use motion imitation and adopt a real-world fine-tuning approach rather than learning completely from scratch. Lastly, rather than manually resetting the robot or hand-designing a recovery controller specific to the robot, we used RL to automatically produce a reset controller, an approach that can be applied to automatically produce reset controllers for other quadrupeds as well.
\begin{figure*}[t]
\begin{minipage}[t]{0.65\textwidth}
\vspace{12pt}
    \centering
    \includegraphics[width=.8\linewidth]{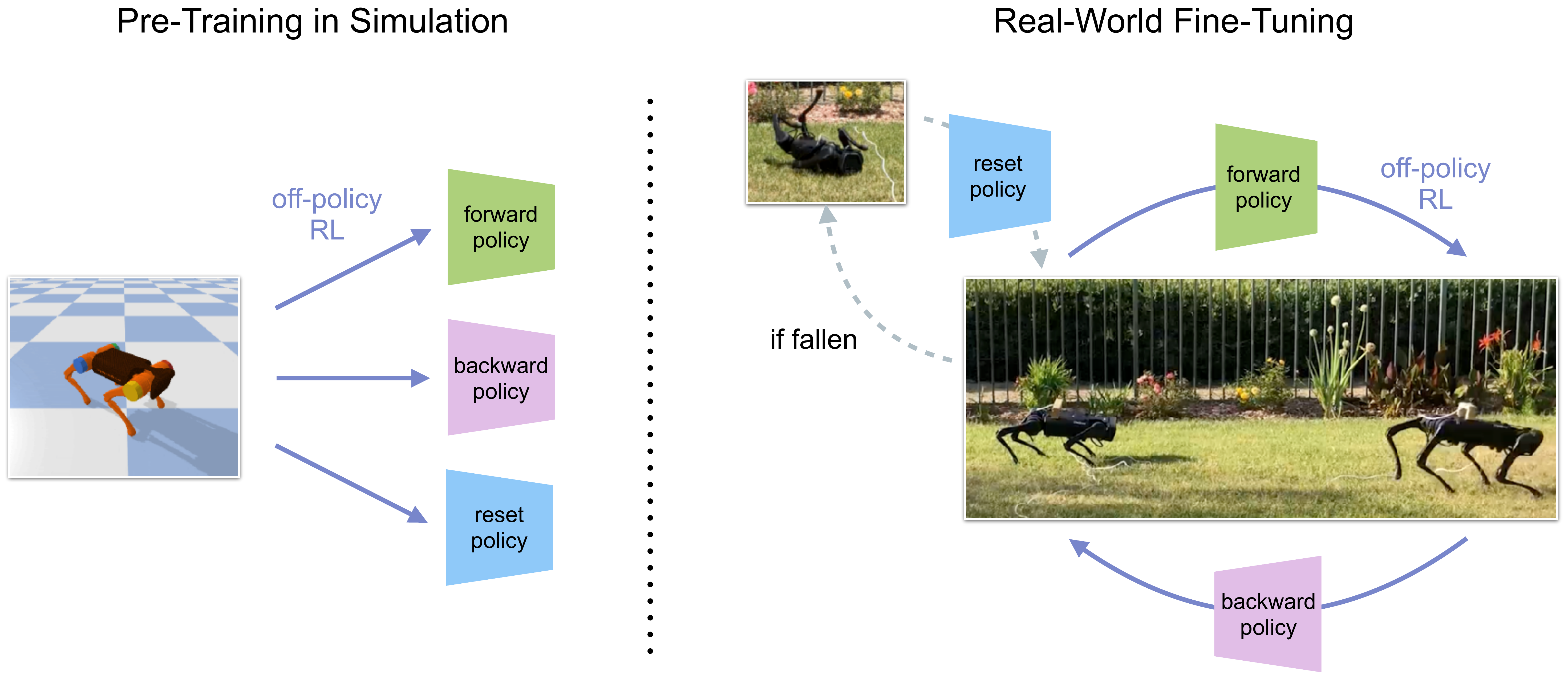}
    \caption{\footnotesize 
    Example of our system. First, we pre-train skills (in this example, forward/backward pacing and reset) in simulation using RL. We then deploy the policies in the real world. The robot executes forward or backward pacing depending on which will bring it closer to the origin. After each episode, it automatically runs its reset policy in preparation for the next. We continue to update the policies with the data collected in the real world using the same RL method to facilitate perpetual improvement.
    }
    \label{fig:overview}
\end{minipage}
\noindent
\hfill
\begin{minipage}[t]{0.32\textwidth}
\vspace{0pt}
\begin{algorithm}[H]
  	\caption{\footnotesize \texttt{TRAIN}: RL Subroutine}
  	\label{alg:rl_update}
  	\begin{algorithmic}[1]{
    \footnotesize
  	\Require Critic to actor update ratio $K$
  	\Require $\mathcal{M}_i, Q_{\theta_i}, \policy_i, \buff_i$
    \item[]
  	\State {{\textsc{// Collect data}}}
  	    \State Calculate the goal $\goal_t$ from $\mathcal{M}_i$.
      	\State Collect trajectory $\traj$ with $\policy_i(\action \mid \state, \goal)$.
      	\State Store $\traj$ in $\buff_i$.
    \Statex
  	\State {{\textsc{// Perform updates}}}
    \For{iteration $i=1, 2, \dots, n_\text{updates}$}
      	\State Update $Q_{\theta_i}$ by minimizing $\mathcal{L}^{\tt REDQ}_{\tt critic}$.
      	\If{i \% $K == 0$}
  	        \State Update $\policy_i$ by minimizing $\mathcal{L}^{\tt REDQ}_{\tt actor}$.
  	    \EndIf
     \EndFor
    \item[]
      $\textbf{return} \ \mathcal{M}_i, Q_{\theta_i}, \policy_i, \buff_i$
     }
  	\end{algorithmic}
\end{algorithm}
\end{minipage}
\vspace{-.25cm}
\end{figure*}

\section{Fine-tuning Locomotion in the Real World}
\label{sec:method}

In this section, we present our system for fine-tuning locomotion policies. It combines a stable and efficient RL algorithm with multi-task training, thereby allowing our robot to learn quickly with minimal human intervention.
Sample-efficient learning is achieved by using a recently proposed off-policy RL algorithm, randomized ensembled double Q-learning (REDQ), which has demonstrated efficient learning in simulated environments \cite{Chen2021RandomizedED}. To enable autonomous training in the real world without requiring human intervention, we stitch together episodes with a learned reset policy.
\paragraph{Overview} Our framework, shown in~\autoref{fig:overview}, involves learning a set of policies, one for each desired skill. 
Because RL algorithms are data intensive and untrained policies can be dangerous on a physical robot, we pre-train our policies in simulation, as is typical for legged locomotion controllers~\cite{Tan2018SimtoRealLA, Lee2020LearningQL, Siekmann2021BlindBS}.
In this phase, we independently train a policy $\pi_i$ for each of the skills, including a recovery policy.
Once the policies are pre-trained in simulation, we perform fine-tuning in the real world by simply continuing the training process using the same RL algorithm. Because the dynamics may be significantly different in the real world, we reset the replay buffers $\mathcal{D}$ for each policy. 
After each episode, the learned recovery policy resets the robot in preparation for the next rollout, preventing time- and labor-intensive manual resets.
For some skills, we use a multi-task framework, which leverages multiple skills to further facilitate autonomous learning.
Algorithm \ref{alg:main} provides an overview of the complete training process.

\paragraph{Motion imitation}
Our policies are trained to perform different skills by imitating reference motion clips using the framework proposed by~\cite{Peng2020LearningAR}. Given a reference motion $\mathcal{M}$ comprising a sequence of poses, a policy is trained to imitate the motion using a reward function that encourages tracking the target poses at each timestep (see~\autoref{sec:mi_reward_details}). This general framework allows us to learn different skills by simply swapping out the reference motion. 
We also learn a recovery policy within this framework by training the robot to imitate a standing pose along with a few important modifications; see~\autoref{sec:training_procedure} for details. 

\begin{algorithm}
  	\caption{\footnotesize Real-World Fine-Tuning with Pre-training}
  	\label{alg:main}
  	\begin{algorithmic}[1]{
\footnotesize
  	 \Require $\nrefs$ reference motions $\{\mathcal{M}_i\}_{i=1}^\nrefs$. 
  	 \State Initialize: Q-function ensembles $\{Q_{\theta i}\}_{i=1}^{\nrefs+1}$, policies $\{\policy_i\}_{i=1}^{\nrefs+1}$, replay buffers $\{\buff_i\}_{i=1}^{\nrefs+1}$, skills $\mathcal{S}$.
    \item[]
  	\State {{\textsc{// Pre-training in simulation}}} 
  	\For{skill $i=0, 2, \dots, \nrefs$}
        \Repeat
          	\State $S_i \leftarrow \texttt{TRAIN}(\mathcal{M}_i, Q_{\theta_i}, \policy_i, \buff_i)$
        \Until{convergence}
      	\State $\mathcal{S} \leftarrow \mathcal{S} \cup \{S_i\}$
      \EndFor

    \item[]
  	\State {{\textsc{// Real-world fine-tuning}}}
  	\Repeat
  	    \State Choose a skill $S_i$ to execute.
  	    \If{in new environment}
  	        \State Optionally clear $\buff_i$
  	    \EndIf
  	    \State $S_i \leftarrow \texttt{TRAIN}(\mathcal{M}_i, Q_{\theta_i}, \policy_i, \buff_i)$
  	\Until{forever}
  	}
  	\end{algorithmic}
\end{algorithm}

\paragraph{Off-policy RL} We leverage the off-policy REDQ algorithm~\cite{Chen2021RandomizedED}, a simple extension to SAC~\cite{Haarnoja2018SoftAO} that allows for a larger ratio of gradient steps to time steps, for sample-efficient RL. REDQ utilizes an ensemble of $Q$-functions $Q_\theta = \{Q_{\theta^k}\}_{k=1}^{N_\text{ensemble}}$ that are all trained with respect to the same target value, which is computed by minimizing over a random subset of the ensemble. This avoids overestimation issues that can occur when using too many gradient steps. 
Our update procedure is summarized in Algorithm~\ref{alg:rl_update}.

\section{System Design}
\label{sec:system}

We use the A1 robot from Unitree as our robot platform and build our simulation using PyBullet~\cite{coumans2021}. For our motion imitation skills, we retarget a mocap recording of dog pacing from a public dataset~\cite{Zhang2018ModeadaptiveNN} and an artist generated side-step motion for the A1 using inverse-kinematics (see~\cite{Peng2020LearningAR}). 
The policies $\{\policy_i\}_{i=1}^\nrefs$ and Q-functions $\{Q_{\theta_i}\}_{i=1}^\nrefs$ are modeled using separate fully-connected neural networks. Updates are computed using the Adam optimizer~\cite{Kingma2017Adam} with a learning rate of $10^{-4}$ and a batch size of 256 transitions. All networks are constructed and trained using TensorFlow~\cite{tensorflow2015-whitepaper}.

\subsection{State and Action Spaces} 
The state $\state_t$ contains a history of 3 timesteps for each of the following features: root orientation (read from the IMU), joint angles, and previous actions. Similar to prior motion imitation approaches~\cite{Peng2018DeepMimicED, Peng2020LearningAR}, the policy receives not only proprioceptive input but a goal $\goal_t$, which comprises the target poses (root position, root rotation, and joint angles) calculated from the reference motion for future timesteps.
In our experiments, we use 4 future target poses, the latest of which is a target for approximately 1 second ahead of the current timestep. 
Actions $\action_t$ are PD position targets for each of the 12 joints and applied at a frequency of 33Hz. To ensure smoothness of the motions, we process the PD targets with a low-pass filter before supplying them to the robot.

\begin{figure}[t]
\vspace{5pt}
    \centering
    \includegraphics[width=.8\linewidth]{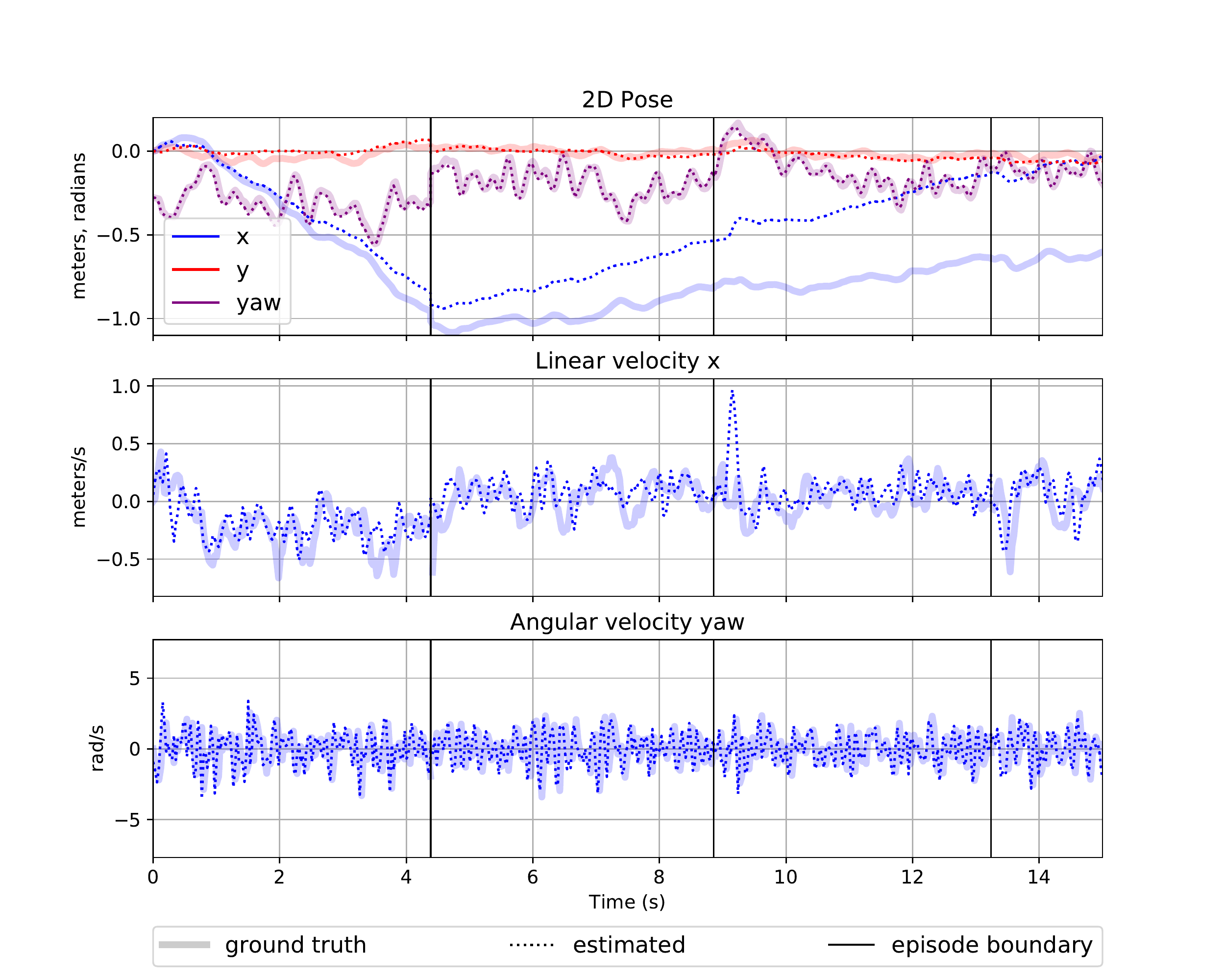}
    \caption{\footnotesize 
    Real-world state estimation compared to motion capture for a robot walking indoors. Yaw and yaw velocity are very accurate, linear velocity is acceptable, and x- and y-position drift over the course of multiple episodes.
    }
    \label{fig:state_estimation}
\vspace{-0.2cm} 
\end{figure}

\subsection{Reward Function}
\label{sec:mi_reward_details}
We adopt the reward function from \cite{Peng2020LearningAR}, where the reward $r_t$ at each timestep is calculated according to:
\begin{equation}
    r_t = w^\text{p} r^\text{p}_t + w^\text{v} r^\text{v}_t + w^\text{e} r^\text{e}_t + w^\text{rp} r^\text{rp}_t  + w^\text{rv} r^\text{rv}_t
\end{equation}
\[
    w^\text{p}=0.5, \ w^\text{v}=0.05, \ w^\text{e}=0.2, \ w^\text{rp}=0.15, \ w^\text{rv}=0.1
\]
The pose reward $r^\text{p}_t$ encourages the robot to match its joint rotations with those of the reference motion.
Below, $\hat{q}_t^j$ represents the local rotation of joint $j$ from the reference motion at time $t$, and $q_t^j$ represents the robot's joint,
\begin{align}
    r^\text{p}_t &= \mathrm{exp}\left[ -5 \sum_j ||\hat{q}_t^j - q_t^j ||^2\right].
\end{align}
$r^\text{v}_t$ and $r^\text{e}_t$ assume a similar form but encourage matching the joint velocities and end-effector positions, respectively. Finally, the root pose reward $r^\text{rp}_t$ and root velocity reward $r^\text{rv}_t$ encourage the robot to track the reference root motion. 
See~\cite{Peng2020LearningAR} for a detailed description of the reward function. 

To estimate the linear root velocity during real-world training, we use a Kalman filter that takes acceleration and orientation readings from the IMU, then corrects them with the foot contact sensors. When a sensor is triggered, we take that foot as a point of 0 velocity, calculate the root velocity using the leg joint velocities, and correct the estimate from the IMU.
We integrate the linear velocity for a rough approximation of the robot's position. Some example data is shown in Figure \ref{fig:state_estimation}. We find that the angular velocity and orientation readings are very accurate, linear velocity is reasonable, and position drifts but is good enough within each episode for our reward calculations.

\subsection{Reset Controller}
\label{sec:training_procedure}

\begin{figure}[t]
\vspace{8pt}
        \centering
        \includegraphics[width=\linewidth]{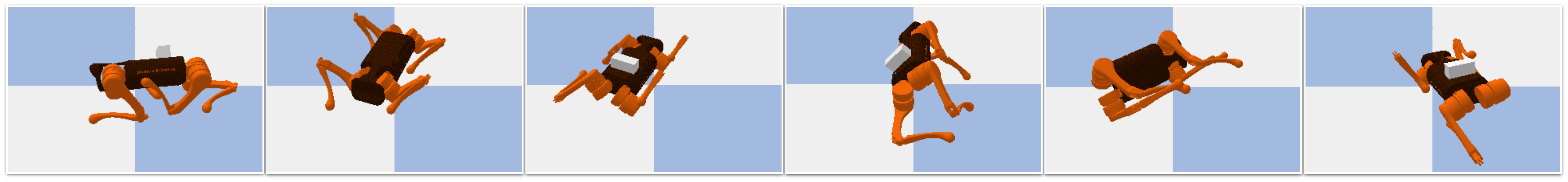}
        \caption{
        \footnotesize Examples from the initial state distribution used to train the reset policy in simulation. To get these states, we drop the robot from about a half meter above the ground with a random initial orientation of the robot's torso. Specifically, its roll, pitch and yaw are drawn from uniform distributions over the intervals $[-\frac{3\pi}{4}, \frac{3\pi}{4}], [-\frac{\pi}{4}, \frac{\pi}{4}]$, and $[-\pi, \pi]$, respectively.
        }
        \label{fig:recovery_dropped_states}
        \vspace{-.1cm} 
\end{figure}
Similar to \cite{Lee2019RobustRC}, the reset policy is trained in simulation by generating a diverse set of initial states. At the start of each episode, we drop the robot from a random height and orientation (see \autoref{fig:recovery_dropped_states}). The robot's objective then is to recover back to a default standing pose. \cite{Lee2019RobustRC} stitches together two policies for reset, first, a self-righting behavior which puts the robot in a stable sitting position, followed by a stand-up behavior which then puts the robot in a standing pose. 

We find that we are able to train a single, streamlined reset policy by modifying the motion imitation objective. Rather than using a reference motion to prescribe exactly how the robot should stand up, we modify our standard imitation reward as follows. First, the policy is only rewarded for rolling right side up. If the robot is upright, we add the motion imitation reward, where the reference is a standing pose, to encourage the robot to stand. Note that although the objective for this reset policy is simple, the behaviors that it acquires are quite complex and agile. This policy is significantly more versatile than hand-designed recoveries used in prior work~\cite{Ha2020LearningTW} or  company-provided reset motions that take more than 10 seconds --- it can recover quickly from a fall by rolling and jumping upright, and if the robot is already upright, it quickly stabilizes it for the next trial. We encourage the reader to view a video of the agile reset controller on the project website.
In our work, we found that the reset policy transfers successfully to all our test terrains, so we perform no fine-tuning. 

\section{Experiments}
\label{sec:experiments}
We aim to answer the following through our experiments:
    \begin{enumerate}[(1)]
    \item How does our finetuning-based method compare to prior approaches that utilize simulated training, including those that perform real-world adaptation?
    \item What effects do our design decisions have on the feasibility of real-world training?
    \item How much can autonomous, online fine-tuning improve robotic skills in a range of real-world settings?
\end{enumerate}

\subsection{Simulation Experiments}
\label{sec:simulation-experiments}
To compare our approach to prior methods, we evaluate on a simulated transfer scenario, where the policy is first trained in one simulated environment, and then ``deployed'' to another simulated setting, which is meant to be representative of the kind of domain shifts we would expect in real-world deployment. Performing this comparison in simulation allows us to abstract away other parts of the adaptation process, such as resets, since prior methods generally do not handle this, and allows us to provide highly controlled low-variance comparisons for each method.
\begin{figure}[t]
\vspace{5pt}
\centering
  \hspace{.16cm}
  \includegraphics[width=.27\linewidth]{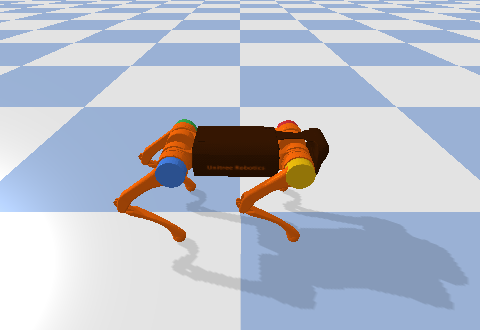} \hspace{.16cm}
  \includegraphics[width=.27\linewidth]{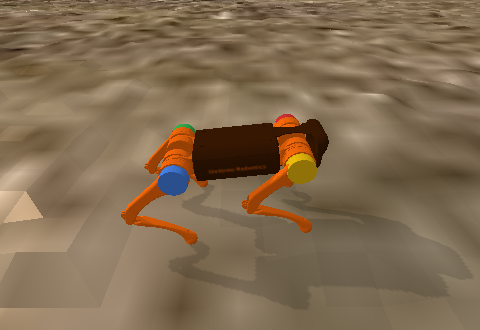} \hspace{.16cm}
  \includegraphics[width=.27\linewidth]{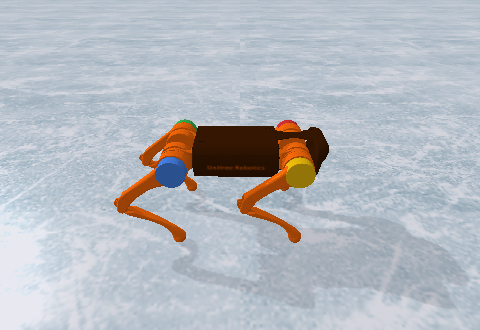}
\\
  \includegraphics[width=.33\linewidth]{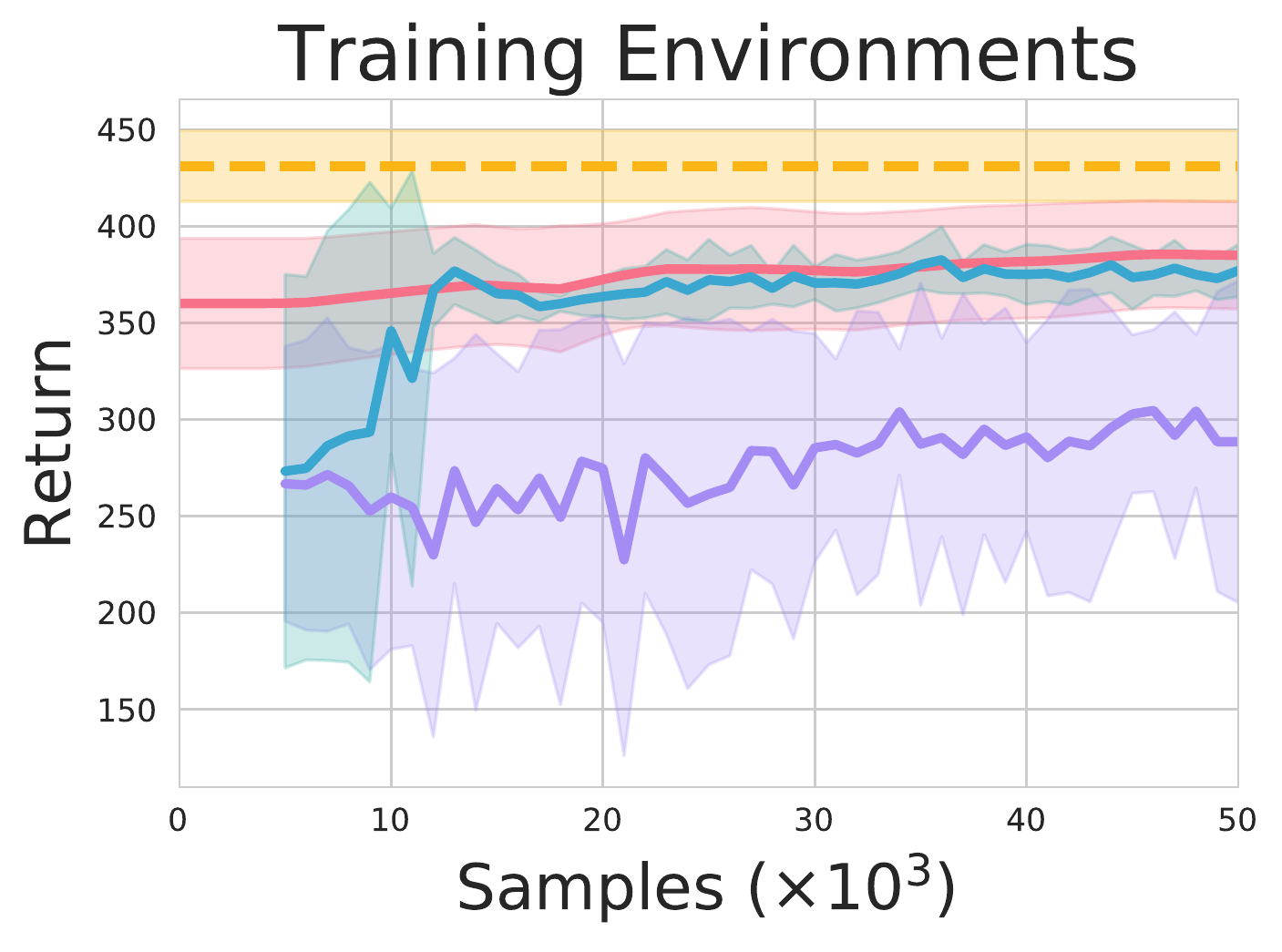}\hspace{-.1cm}
  \includegraphics[width=.33\linewidth]{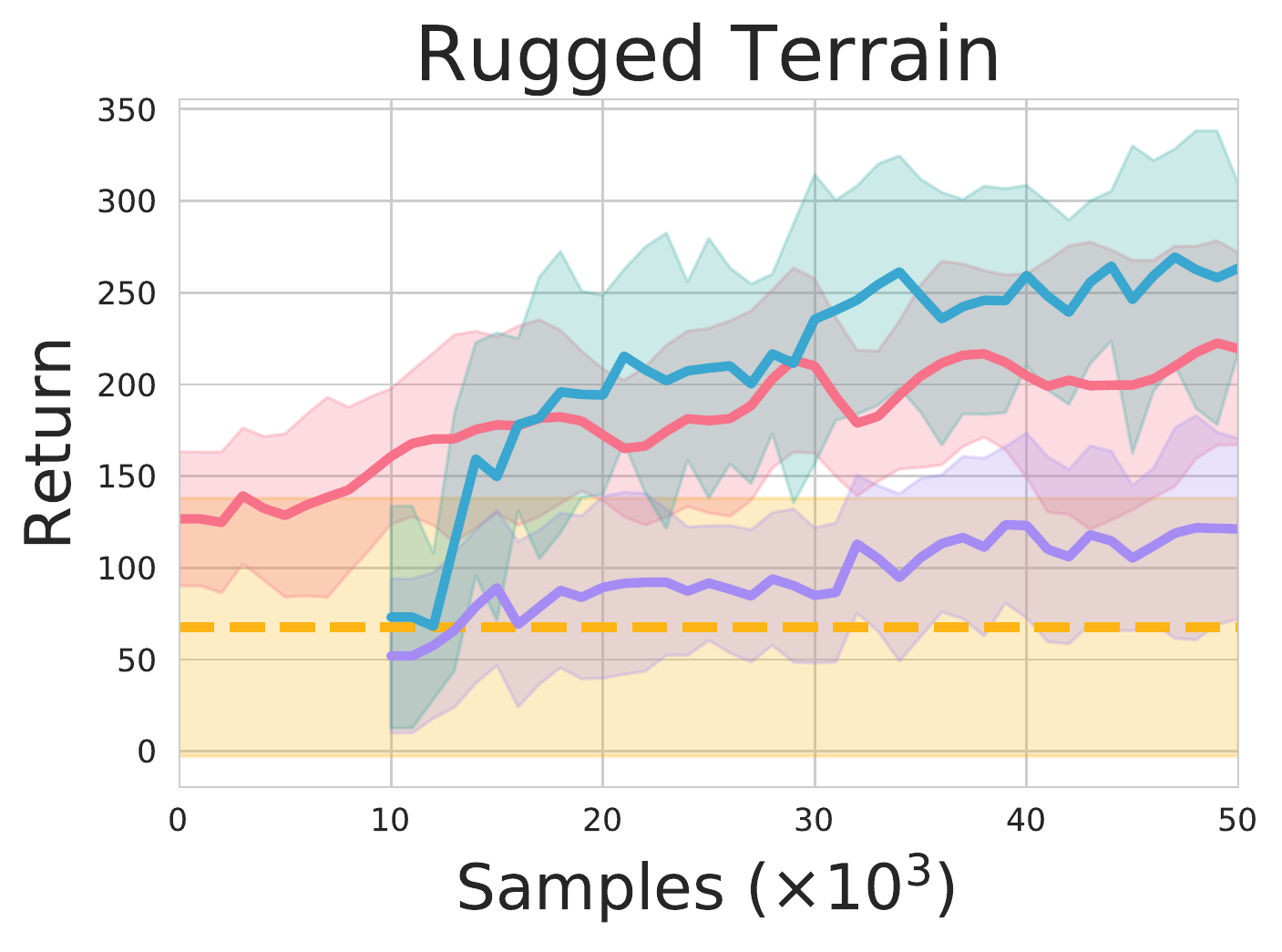}\hspace{-.1cm}
  \includegraphics[width=.33\linewidth]{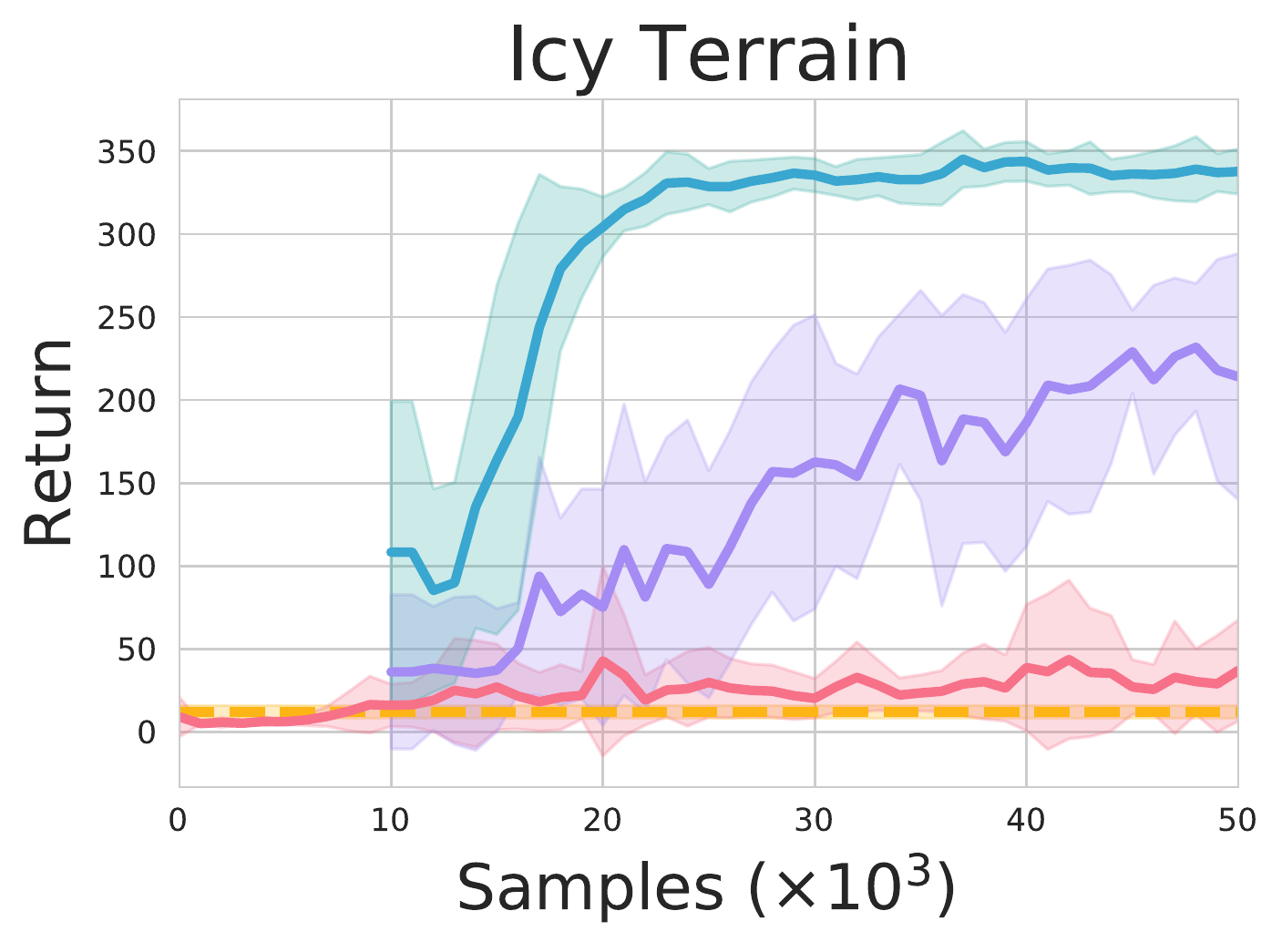}\\
  \vspace{-.1cm}\includegraphics[width=.75\linewidth]{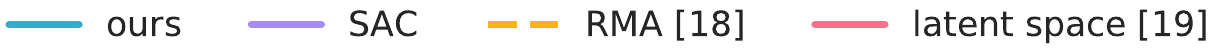}
 \caption{\footnotesize We report each adapted policy's performance (mean and standard deviation across 10 trials) in the target domain (pictured in the top row) with respect to the samples used for adaptation.
 Prior methods that learn adaptation strategies during training (pink and yellow) excel in environments that are similar to those seen during training (left) but fail in environments that are sufficiently different (middle, right). In contrast, our fine-tuning method (blue) continues to improve under all circumstances. We also note that our use of REDQ (blue) improves over SAC (purple).
 }
\label{fig:adaptation-comparison}
\end{figure}
\paragraph{Comparison to prior work}
To address (1), we compare our fine-tuning method to prior work, adapting a learned forward pacing gait to several test environments. To this end, we pre-train all methods with standard dynamics randomization (varying mass, inertia, motor strength, friction, latency) on flat ground until convergence, and then deploy them on various terrains for adaptation. The test terrains include a flat ground terrain that is similar to the environments seen during pre-training, as well as two terrains that differ substantially from the training settings:  randomized heightfield simulating rugged terrain, and a low-friction surface simulating a slippery, icy terrain (with a friction coefficient far below that seen during training). Note that we intentionally select these test settings to be different from the training environment: while prior works generally carefully design the training environments to cover the range of settings seen at test-time, we intentionally want to evaluate the methods under \emph{unexpected} conditions. For each type of environment, we use 10 instantiations of the environment with different dynamics parameters, and test each method's ability to adapt to them. The dynamics of the environment is fixed for each adaptation trial, and the same 10 environments are used for all methods.

Peng et al.~\cite{Peng2020LearningAR} uses dynamics randomization to learn a latent representation of behaviors that are effective for different settings of the dynamics parameters.
During adaptation, this method searches in the latent space for a behavior that maximizes the agent's test-time performance using AWR~\cite{peng2019advantage}. 
RMA also uses dynamics randomization to learn a latent representation of different strategies. But instead of searching in latent space during adaptation, RMA trains an additional `adaptation module,' similarly to Yu et al.~\cite{Yu2019SimtoRealTF, Yu2020LearningFA}, to predict the appropriate latent encoding given a recent history of observations and actions. Since RMA simply corresponds to a policy with memory and does not actually use data collected from a new domain to update the parameters of the model, it does not improve with additional data. We implement both methods using SAC as the policy optimizer at training-time, to match our method.

\paragraph{REDQ vs. SAC} One of our design decisions is to use the recently-proposed REDQ algorithm for policy optimization. To evaluate the importance of this choice, we consider an ablation of our method that uses vanilla SAC instead of REDQ for both pre-training and fine-tuning.
For REDQ, we use 10 Q-functions, and randomly sample 2 when computing the target value. For both fine-tuning methods, we collect an initial buffer of samples before starting to fine-tune (hence these curves start at 5000 or 10000 samples).

We report the adaptation performance of each method in~\autoref{fig:adaptation-comparison}. When tested on the training environments, we see that both RMA and the latent space method perform well. This indicates that these methods indeed excel in regimes where the environments seen at training-time resemble the ones that the method must adapt to at test-time. However, when these policies, which are trained on flat ground, are placed on uneven or extremely slippery terrain, they exhibit a significant drop in performance. Both methods suffer because they assume that a pre-trained encoder or latent space can generalize, which is too strong an assumption when the test environment differs sufficiently from the training environments. RMA especially suffers in this case because it relies entirely on the pre-trained encoder to adapt, and when this encoder fails to generalize, it does not have any other recourse. The latent space method does adapt, but it relies on the latent space already containing suitable strategies for the new environment and ends up with a suboptimal policy. 
In contrast, our finetuning approach is able to continuously improve and eventually succeed.
These results show that pretrained models, even prior adaptive models, can fail if tested on environments that deviate too much from those seen in training. Our method initially fails also, but is able to recover good performance through fine-tuning.

\subsection{Real-World Experiments}
\begin{figure}[t]
\vspace{5pt}
  \includegraphics[width=.95\linewidth]{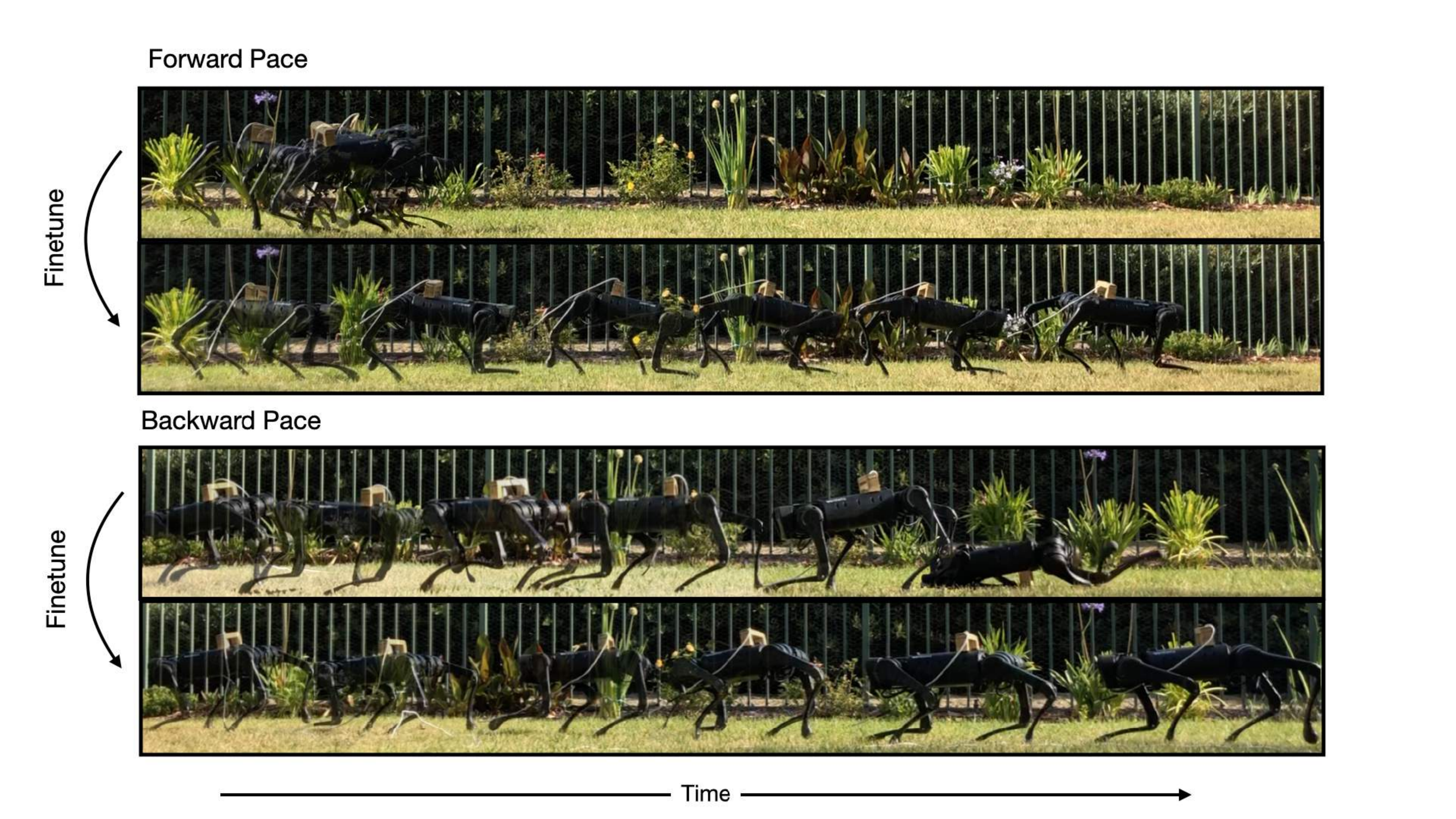}\\
  \includegraphics[width=.955\linewidth]{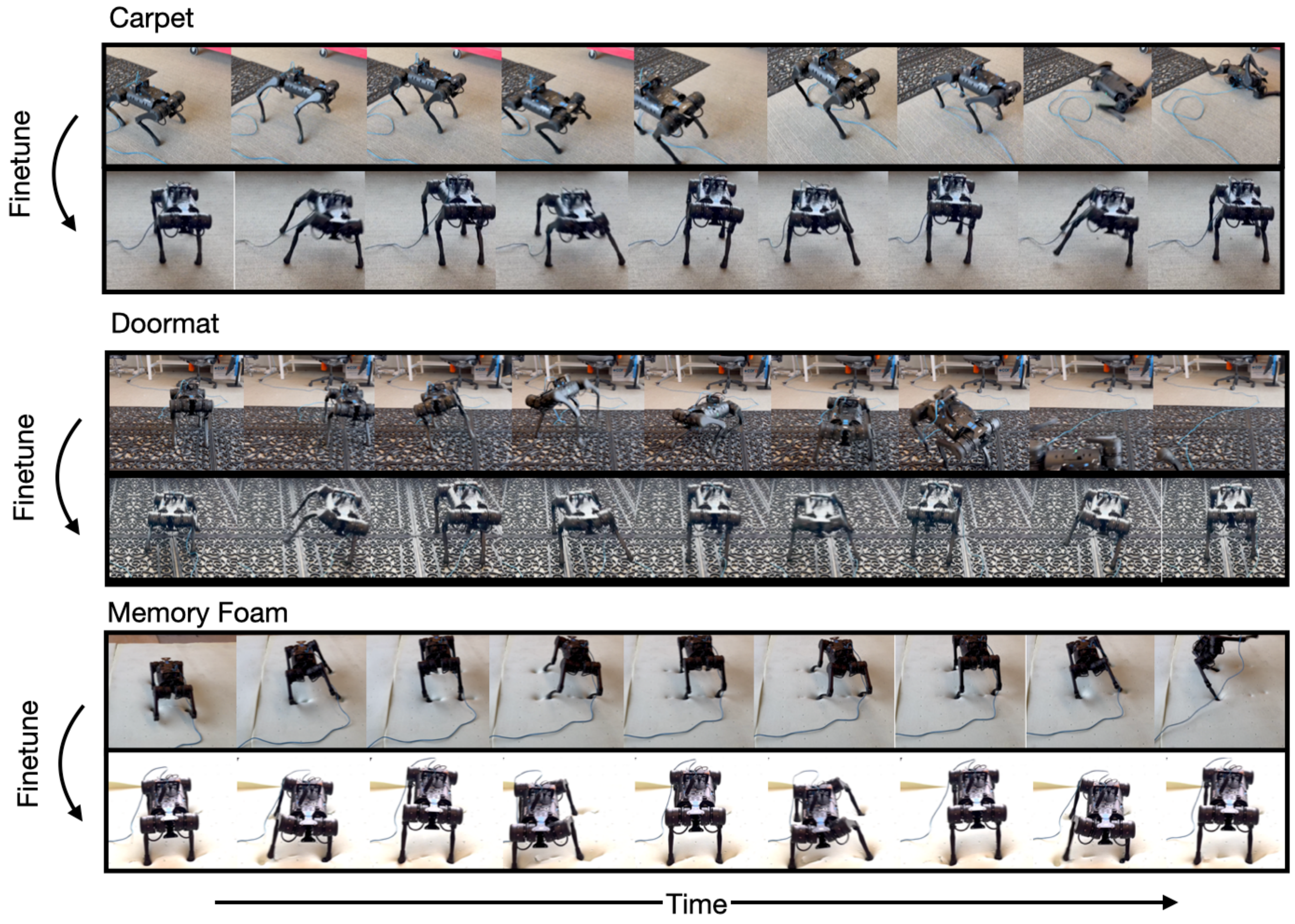}
 \caption{ \footnotesize \textbf{1st and 2nd Rows:} Example rollouts of the pacing policies before and after fine-tuning. The figure depicts a timelapse of a rollout of each policy, where the opacity of the robot indicates time progression (i.e., more opaque corresponds to more recent). Before fine-tuning, the forward policy struggles to make progress, while the pre-trained backwards policy makes fast progress but is quite unstable. After fine-tuning, both policies make forward progress without falling. \textbf{3rd to 5th Rows:} Example rollouts of the side-step policies before and after fine-tuning. On all terrains, the pre-trained policies fail to complete the task without falling. After fine-tuning, the polices are fine-tuned to successfully and reliably execute the skill on all three domains.
 }
\label{fig:pacing-filmstrip}
\end{figure}

Our real-world experiments aim to evaluate how much autonomous online fine-tuning can improve a variety of robotic skills in a range of realistic settings. We evaluate our fine-tuning system in four real-world domains: an outdoor grassy lawn, a carpeted room, a doormat with crevices, and memory foam (see \autoref{fig:pacing-filmstrip}),
each of which presents unique challenges. The outdoor lawn presents a slippery surface, where the feet can either slip on the grass or get stuck in the dirt. In this domain, we finetune a pacing gait in which both legs on one side of the body swing forward in unison. For the indoors experiments, the robot is tasked with performing a side-stepping motion on the various surfaces mentioned above.
The carpeted room, in contrast to the grass, is high-friction, causing the robot's soft rubber feet to deform in a manner inconsistent with simulation.
The doormat presents a textured surface for the feet to get stuck. The $4$cm-thick memory foam mattress is especially difficult, because the feet can sink into the mattress and the gaits need to change substantially to achieve ground clearance.
For all experiments, we pretrain the policy on flat ground in simulation, run it for 5000 samples to initialize the buffer, and then finetune the policy in the real world.

\begin{figure}[t]
\vspace{5pt}
\centering
  \hspace{-.1cm}
  \includegraphics[width=.45\linewidth]{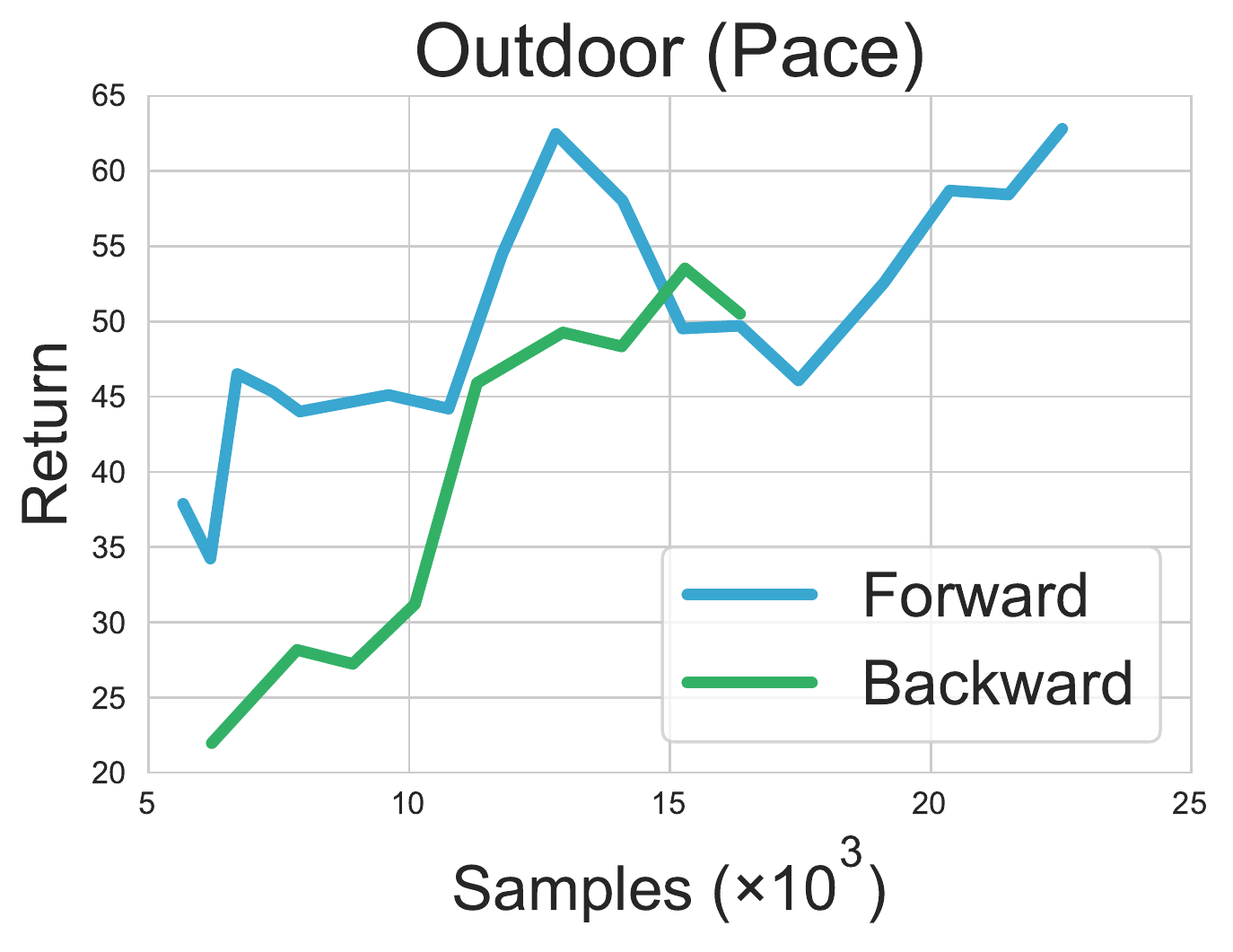}
  \includegraphics[width=.45\linewidth]{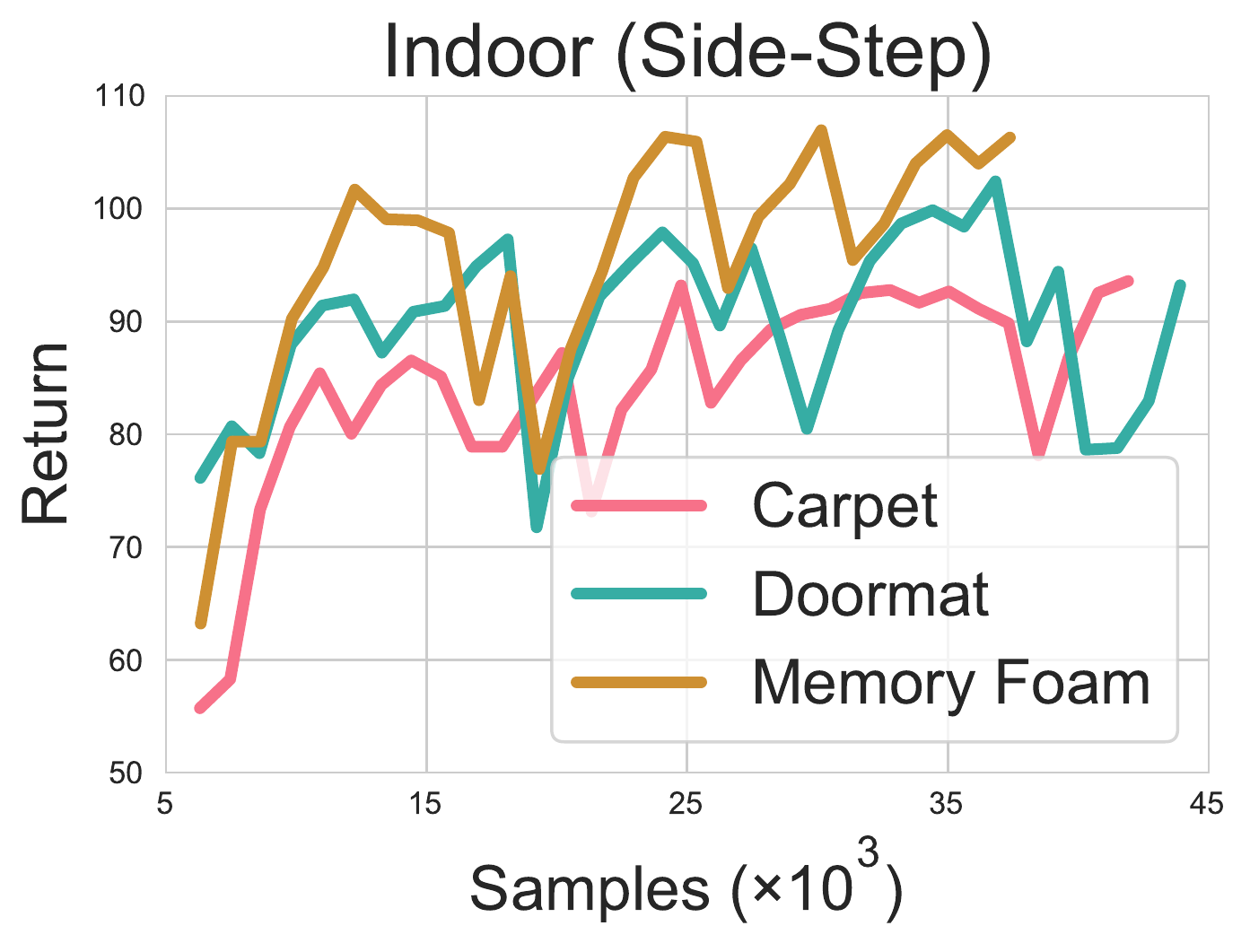}\hspace{-.1cm}
  \\ \vspace{-.05cm} \includegraphics[width=.8\linewidth]{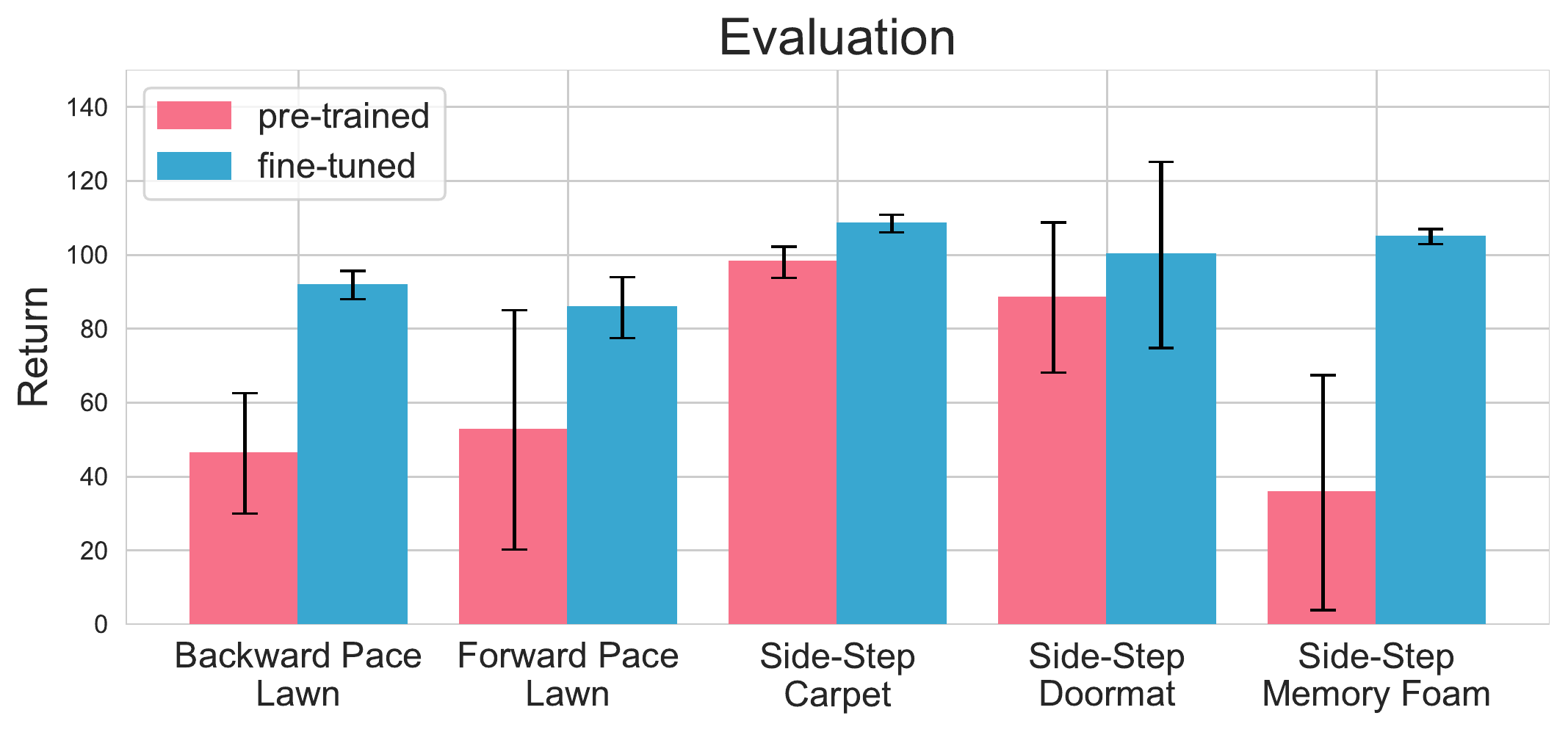}
 \caption{\footnotesize \textbf{Top:} Learning curves for all the real-world fine-tuning experiments showing the average return of data collected by a stochastic policy during each iteration of training. \textbf{Bottom:} We evaluate each policy with a deterministic policy before and after fine-tuning and report mean and standard deviation over ten trials. In all domains, fine-tuning leads to improvement, and the improvement is particularly pronounced on the most difficult surfaces, such as the lawn and memory foam mattress.
 }
\label{fig:fine-tune-plots}
\vspace{-0.2cm}
\end{figure}

We report the average return of the policies during training in~\autoref{fig:fine-tune-plots} with respect to the number of real-world samples collected. In all environments, our framework leads to substantial performance improvement with a modest amount of data. On the lawn, the pre-trained forward pacing policy makes very little forward progress,
whereas the pre-trained backward pacing policy tends to trip and fall. After less than 2 hours in total of operation, the robot learns to consistently and stably pace forward and backward with very few failures. Indoors, the pre-trained sidestepping policy tends to twitch violently and fall to the floor before completing its motion. This is true across all the terrains: carpet, memory foam, and doormat with crevices. But on each terrain, in less than 2.5 hours of training, the robot learns to consistently execute the skill without stumbling. This figure includes overhead such as swapping batteries and handling robot malfunctions. 
\autoref{fig:pacing-filmstrip} shows a comparison of the behaviors of the policy before and after fine-tuning. 
For all experiments, we include videos of the training process as well as evaluation of the performance before and after training on the project website.

\paragraph{Semi-autonomous training} In our experiments, we find that our learned recovery controller is very effective in providing efficient, automatic resets in between trials. Over all experiments, the recovery policy was 100\% successful. We compare our reset controller to the built-in rollover controller from Unitree. On hard surfaces, both controllers are effective while the built-in one is substantially slower than the learned policy. On the memory foam, the built-in controller performs less reliably. Please see the supplementary video or project website for videos of the reset policy in all environments and a comparison to the built-in controller. When the robot occasionally drifts outside the workspace area, it is re-positioned by a human supervisor. The need for this intervention, though, was greatly reduced by the use of the simultaneous learning of complementary skills. 

\section{Conclusion}
\label{sec:conclusion}
We present a system that enables legged robots to fine-tune locomotion policies in real-world settings, such as grass, carpets, doormats and mattresses, via a combination of autonomous data collection and data-efficient model-free RL. Our system provides for automated recoveries from falls, reward calculation through state estimation using onboard sensors, and data-efficient fine-tuning of a variety of locomotion skills.  The fine-tuning improves the performance substantially, reaching a high level of proficiency even when starting with a gait that frequently stumbles and falls. In this work, we focus on fine-tuning in each environment separately. It would be interesting to adapt our system into a lifelong learning process, where a robot never stops learning. When the robot encounters a new environment, the fine-tuning will quickly adapt its policy. When the robot stays in the same environment for an extended period of time, the fine-tuning will gradually perfect its skills. We plan to test such a lifelong learning system for legged robots in complex, diverse and ever-changing real-world environments.

\subsection*{Acknowledgements}
This work was supported by ARL DCIST CRA W911NF-17-2-0181, the Office of Naval Research, and Google. Laura Smith is supported by NSF Graduate Research Fellowship.

\balance
\bibliography{references}
\bibliographystyle{IEEEtran}

\end{document}